\documentclass[lettersize,journal]{IEEEtran}
\usepackage{amsmath,amsfonts}
\usepackage{algorithmic}
\usepackage{algorithm}
\usepackage{array}
\usepackage[caption=false,font=normalsize,labelfont=sf,textfont=sf]{subfig}
\usepackage{textcomp}
\usepackage{stfloats}
\usepackage{url}
\usepackage{verbatim}
\usepackage{graphicx}
\usepackage{cite}
\usepackage{makecell}
\usepackage{multirow}
\begin{document}

\title{Multi-modal Reference Learning for Fine-grained Text-to-Image Retrieval}

\author{
Zehong Ma, 
Hao Chen,
Wei Zeng,
Limin Su,
and Shiliang Zhang,~\IEEEmembership{Senior Member,~IEEE}
\thanks{This work is supported in part by Grant No. 2023-JCJQ-LA-001-088, in part by the Natural Science Foundation of China under Grant No. U20B2052, 61936011, 62236006, 62402013, in part by the China Postdoctoral Science Foundation under Grant No. 2023M730056, in part by the Okawa Foundation Research Award, in part by the Ant Group Research Fund, and in part by the Kunpeng\&Ascend Center of Excellence, Peking University.}
\thanks{Z. Ma, H. Chen, W. Zeng, and S. Zhang are with the State Key Laboratory of Multimedia Information Processing, School of Computer Science, Peking University, Beijing 100871, China. E-mail: zehongma@stu.pku.edu.cn, \{hchen, weizeng, slzhang.jdl\}@pku.edu.cn.}
\thanks{L. Su is with Beijing Union University, Beijing, China. E-mail: xxtlimin@buu.edu.cn}
}

\markboth{IEEE TRANSACTIONS ON MULTIMEDIA}{Ma \MakeLowercase{\textit{et al.}}: Multi-modal Reference Learning for Fine-grained Text-to-Image Retrieval}


\maketitle

\begin{abstract}
Fine-grained text-to-image retrieval aims to retrieve a fine-grained target image with a given text query. Existing methods typically assume that each training image is accurately depicted by its textual descriptions. However, textual descriptions can be ambiguous and fail to depict discriminative visual details in images, leading to inaccurate representation learning. To alleviate the effects of text ambiguity, we propose a Multi-Modal Reference learning framework to learn robust representations. We first propose a multi-modal reference construction module to aggregate all visual and textual details of the same object into a comprehensive multi-modal reference. The multi-modal reference hence facilitates the subsequent representation learning and retrieval similarity computation. Specifically, a reference-guided representation learning module is proposed to use multi-modal references to learn more accurate visual and textual representations. Additionally, we introduce a reference-based refinement method that employs the object references to compute a reference-based similarity that refines the initial retrieval results. Extensive experiments are conducted on five fine-grained text-to-image retrieval datasets for different text-to-image retrieval tasks. The proposed method has achieved superior performance over state-of-the-art methods. For instance, on the text-to-person image retrieval dataset RSTPReid, our method achieves the Rank1 accuracy of 56.2\%, surpassing the recent CFine by 5.6\%.
\end{abstract}


\section{Introduction}
\IEEEPARstart{F}{ine-grained} text-to-image retrieval~\cite{li2017person,zhang2018deep} aims to retrieve a fine-grained object of interest from a large-scale image gallery using a text query. Thanks to the advance of deep learning algorithms, recent years have witnessed remarkable progress in fine-grained image classification~\cite{mao2024robust} and retrieval~\cite{9318538}. Compared with image queries~\cite{S2-Net} and multi-modal composed queries~\cite{huang2024attribute}, text queries are more flexible and easier to acquire. Therefore, fine-grained text-to-image retrieval is attracting increasing attention in the research community. 

As a cross-modality retrieval task, fine-grained text-to-image retrieval seeks to bridge the gap between visual and textual modalities by aligning their representations. Following pioneering efforts~\cite{li2017person}, numerous frameworks have emerged to tackle this challenge. For instance, the authors of~\cite{zheng2020dual, zhang2018deep} propose dual-encoder models with innovative loss functions to bridge the gap between visual and textual modalities. Besides, various single-modal or cross-modal attention mechanisms and interaction modules have been proposed in~\cite{chen2018improving,niu2020improving,wang2020vitaa,gao2021contextual,gao2021tvfr,ding2021semantically,shu2023see,yan2022cfine, selfalign,9200698} to extract discriminative representations and narrow the modality gap.

\begin{figure}[t]
	\begin{center}
		\includegraphics[width=1\linewidth]{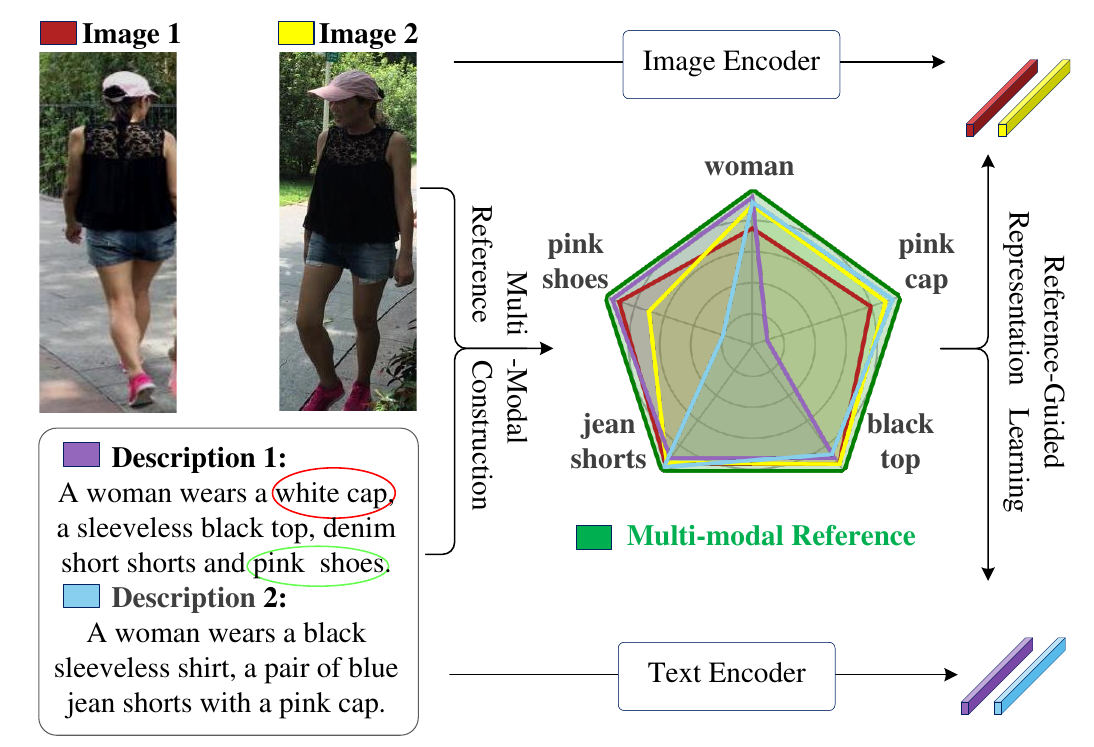}
	\end{center}
	\caption{Illustration of textual ambiguity and our motivation. Two images of the same person and their textual descriptions are illustrated, where the red ellipse shows an inaccurate annotation and the green ellipse highlights the discriminative detail that is missed in the second textual description. Our motivation is to construct a comprehensive multi-modal reference that encompasses all the details of a target object to guide learning better visual and textual representations.
    }
	\label{introduction}
\end{figure}

Existing cross-modal alignment methods usually assume that each training image is accurately depicted by its textual descriptions. However, due to viewpoint variance, occlusion in real-world scenarios, and the subjectivity of annotators, textual descriptions can be inaccurate or incomplete. As illustrated in Fig.~\ref{introduction}, the ``white cap" in the first description is an inaccurate annotation of the pink cap, and the discriminative ``pink shoes" cue is not depicted in the second sentence. Simply considering such ambiguous textual descriptions as ground truth can be harmful to learning accurate visual and textual representations. 

To address the issue of inaccurate representation learning caused by textual ambiguity, we propose a Multi-Modal Reference (MMRef) learning framework, which consists of Multi-Modal Reference Construction (MMRC) and Reference-Guided Representation Learning (RGRL). In MMRC, we first randomly initialize a learnable multi-modal reference embedding for each object. Then, we aggregate all the image and text details of the object into the reference embedding using a global fusion module and a local reconstruction module. 
As a comprehensive representation of the target object, the aggregated multi-modal reference is hence employed to guide learning better visual and textual representations in RGRL. 

The global fusion module is designed to construct a robust multi-modal reference for each target object. The reference adaptively aggregates discriminative visual cues from multiple global visual or textual features of the same object using contrastive learning. It optimizes the reference at a global level, which may ignore some important local cues of the object.
We further introduce a local reconstruction module. This module takes the multi-modal reference as a condition to reconstruct masked textual words so that the reference can encompass more local details. The constructed multi-modal reference contains comprehensive details of the object, which, in turn, guides learning better uni-modal representation to mitigate the effects of ambiguous texts. 

Moreover, the multi-modal references also facilitate the inference stage. We propose a reference-based refinement method, which takes the multi-modal references as an intermediate bridge to narrow the gap between visual and textual modalities.
Taking multi-modal references as shared semantic prototypes, we first project the visual and textual features into a unified reference space. This projection preserves the modality-agnostic semantics while discarding modality-specific details, such as visual backgrounds and textual function words. A reference-based similarity between projected features is computed to augment the initial image-text correlation and refine the retrieval result. 

Extensive experiments are conducted on five fine-grained text-to-image retrieval datasets of three tasks, i.e., CUHK-PEDES~\cite{li2017person}, ICFG-PEDES~\cite{ding2021semantically}, RSTPReid~\cite{zhu2021dssl} for text-to-person retrieval, CUB~\cite{reed2016learning} for text-to-bird retrieval, and Flowers~\cite{reed2016learning} for text-to-flower retrieval, respectively. Our method achieves superior performances over state-of-the-art methods on these tasks. In addition, our experiments on various image-based person re-identification datasets suggest that aligning visual representations with textual descriptions effectively enhances domain generalization ability. 

Our contributions can be summarized as follows. 1) We present a multi-modal reference learning framework named MMRef to enhance uni-modal representations from noisy inputs. A multi-modal reference is first constructed with MMRC, which, in turn, guides learning better uni-modal representations using RGRL. 2) We further propose a test-time reference-based refinement method that leverages multi-modal references as semantic prototypes to compute reference-based similarity to refine the retrieval results. 3) Extensive experiments demonstrate the superiority of our method in five fine-grained text-to-image retrieval tasks. To the best of our knowledge, this is an initial attempt that aligns images with text descriptions to boost the domain generalization capability of visual features for image-based person retrieval.

\section{Related Work}
\label{sec: related_work}
Fine-grained image retrieval can be summarized into four categories of image-to-image, attribute-to-image, text-to-image, and composed image retrieval, respectively, according to the query type. In image-to-image retrieval, S$^2$-Net~\cite{S2-Net} proposes a novel attention branch to learn the human semantic partition to effectively avoid misalignment introduced by even partitioning. Recent PromptSG~\cite{yang2024pedestrian} utilizes prompt-driven semantic guidance from CLIP to extract better visual features. In composed image retrieval, a novel attribute-guided pedestrian retrieval (AGPR)~\cite{huang2024attribute} is introduced to integrate specified attributes with query images to refine retrieval results. Compared with image queries~\cite{S2-Net,yang2024pedestrian}, attribute queries, and composed queries~\cite{huang2024attribute}, text queries are more flexible and easier to acquire. This work is closely related to fine-grained text-to-image retrieval, proxy learning, and domain generalizable person retrieval. This section briefly reviews recent works and discusses our differences with them.

\subsection{Fine-grained Text-to-Image Retrieval}
The challenges of fine-grained text-to-image retrieval lie in extracting discriminative features of images and text, and establishing their cross-modal associations. According to the scale of features used for cross-modal alignment, existing works can be divided into two categories: single-scale and multi-scale representation learning methods. 

Single-scale representation learning methods only take textual or visual features at a unique scale as input to conduct cross-modal alignment. The pioneering work~\cite{li2017person} proposes a benchmark model GNA-RNN that uses word-level token features and global visual features to compute similarities between sentences and images. Based on this work, Li et al.~\cite{li2017identity} first take identity information into account and propose a two-stage framework. In order to project global visual and textual features into a unified space, Zhang et al.~\cite{zhang2018deep} and Zheng et al.~\cite{zheng2020dual} construct end-to-end dual-encoder models and propose dedicated losses. Sarafianos et al.~\cite{sarafianos2019adversarial} leverage adversarial learning to generate global modality-invariant features. Recently, Wang et al.~\cite{wang2022caibc} introduce two separate visual encoders to extract color-dependent and color-independent features individually at the global level and achieve state-of-the-art performance. Li et al.~\cite{li2023your} discuss the issue of false negatives in textual annotations and propose the innovative False Negative Elimination (FNE) method to robustly enhance image-text matching. This method substantially boosts the performance of fine-grained cross-modal retrieval.

Multi-scale representation learning methods use multi-scale textual or visual features to align different modalities. Chen et al.~\cite{chen2018improving} propose GLA method to align global-global and local-local features, while the MIA method in \cite{niu2020improving} further introduces the global-local relationship. 
Guan et al.~\cite{guan2021learning} introduce the CCA-ResNet, a state-of-the-art method in learning nuanced visual features through joint multi-modal training.
A lot of subsequent efforts~\cite{wang2020vitaa,gao2021contextual,ding2021semantically,chen2022tipcb, shu2023see, yan2022cfine} work on introducing more effective interaction or attention modules to acquire the relationship between multi-scale features either explicitly or implicitly. Recent methods \cite{shu2023see,yan2022cfine,he2023vgsg} replace ResNet-50 with the vision transformer to further improve the retrieval performance. Bin et al.~\cite{bin2023unifying} utilize hierarchical alignment transformers to adeptly explore multi-level correspondences of different layers between images and texts, providing a powerful framework for cross-modal retrieval. 

However, existing methods rely on the assumption that textual descriptions of the given image are always complete and accurate, which is unrealistic in real scenarios. Our MMRef introduces a multi-modal reference for each object to alleviate the effects of low-quality textual descriptions.

\subsection{Proxy Learning}
The proxy-based metric learning is first proposed to reduce the training complexity and accelerate training convergence. Meanwhile, the proxy, which can also be called a reference, could effectively alleviate the negative impacts of noisy labels and facilitate learning better representations. 

In the field of computer vision, ProxyNCA~\cite{movshovitz2017no} leverages the proxy, a group of learnable representations, to compare data samples via the neighborhood component analysis (NCA) loss~\cite{roweis2004neighbourhood}. The motivation is to set image samples as anchors to compare with class proxies instead of class samples to reduce sampling times. ProxyNCA++~\cite{teh2020proxynca++} further improves ProxyNCA by scaling the gradient of proxies. Zhu et al.~\cite{zhu2020fewer} propose to sample the most informative negative proxies to improve performance, while Kim et al.~\cite{kim2020proxy} set the proxies as anchors instead of the samples to learn the inter-class structure. Yang et al.~\cite{yang2022hierarchical} develop a hierarchical-based proxy loss to boost learning efficiency. Roth et al.~\cite{roth2022non} regulate the distribution of samples around the proxies following a non-isotropic distribution. These methods focus on learning uni-modal visual proxies for metric learning. 
Recently, some works have utilized the visual proxy in vision-language tasks. Xue et al.~\cite{xue2023clipvip} use a learnable query to aggregate video frames into one visual proxy token. Qian et al.~\cite{qian2023inmap} learn a uni-modal visual proxy based on the predefined textual proxy and unlabeled target images. These methods still focus on constructing a pure visual proxy.

In contrast to those methods that learn a visual proxy in vision or vision-language tasks, our method is an initial attempt to learn a multi-modal proxy from multi-modal inputs. 

\subsection{Domain Generalizable Person Retrieval}
Most of existing methods in image-based person retrieval encounter performance degradation when applied to novel domains due to domain gaps. To address this issue, a domain generalizable retrieval method is proposed in \cite{song2019generalizable}, which aims to train a fine-grained person retrieval model that can work well on unseen domains. Several works \cite{pan2018ibnnet,jin2020style,xuan2022intra, 9330805} try to use instance normalization or consistency to alleviate the degradation caused by domain gaps. Recently, vision-language pre-training models~\cite{clip_radford2021learning, yao2021filip, yang2024pedestrian} have shown an impressive domain generalization capability on various vision tasks. 
This inspires us to explore whether the text descriptions can improve the domain generalization ability of visual representations after training on vision-language datasets. Our experiments show that textual descriptions facilitate learning background-invariant features to improve the domain generalization ability of a retrieval model.

\section{Proposed Method}\label{Methodology}
\begin{figure*}[t]
	\begin{center}
		\includegraphics[width=1\linewidth]{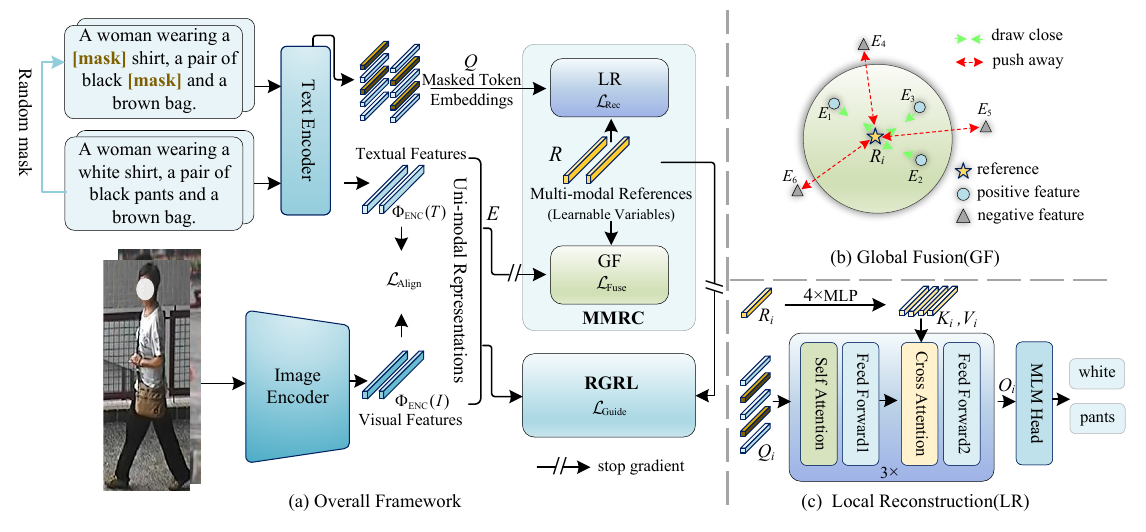}
	\end{center}
        \vspace{-5pt}
	\caption{
		 (a) Overview of MMRef framework. The loss $\mathcal{L}_{\text{Align}}$ is used to align global visual features and textual features. The multi-modal references are constructed in the multi-modal reference construction (MMRC) with a global fusion (GF) module and a local reconstruction (LR) module. In reference-guided representation learning (RGRL), multi-modal references are utilized to facilitate learning better uni-modal representations. (b) The illustration of global fusion for a single reference. (c) The pipeline of local reconstruction for one reference. } 
	\label{Architecture}
\end{figure*}

\subsection{Overview}\label{formulation}
This work aims to retrieve fine-grained images based on text queries. Given a set of text queries $T=\{t_i\}_{i=1}^n$ and a corresponding image gallery $I=\{v_i\}_{i=1}^n$ with a size $n$, our goal is to train a model $\Phi_{\textsc{ENC}}$ that can extract a feature vector for each text query to retrieve an image in gallery $I$ that contains the target object.
We thus define the objective of text-to-image retrieval as seeking the image $g^*$ in $I$ that has the maximal cosine similarity with the text query $q$, which can be defined as:
\begin{equation}
    g^* = \arg\max_{g\in I}\cos(\Phi_{\textsc{ENC}}(q),\Phi_{\textsc{ENC}}(g)),
\end{equation}
where $\Phi_{\textsc{ENC}}(q)$ and $\Phi_{\textsc{ENC}}(g)$ are L2-normalized query and gallery features, and $\cos(\cdot, \cdot)$ denotes the cosine similarity.

Following previous works~\cite{wang2020vitaa, han2021textreid}, we first align the original visual features and textual features extracted by the model $\Phi_{\textsc{ENC}}$ with a contrastive loss. 
The contrastive loss function $\mathcal{L}_{\text{CL}}(S^+, S^-)$ is designed to maximize similarities $S^+$ between positive pairs while minimizing similarities $S^-$ between negative ones, which is defined as follows:
\begin{equation}
\begin{aligned}
    \mathcal{L}_{\text{CL}}(S^+, S^-) = &\sum_{s^p\in S^+}\log[1+e^{-\tau_{p}(s^p -\alpha)}]
    \\& + \sum_{s^n\in S^-}\log[1+e^{\tau_{n}(s^n -\beta)}],
\end{aligned}
\label{loss_func}
\end{equation}
where $\tau_p$ and $\tau_n$ are temperature parameters, $\alpha$ is the lower bound for positive similarity, and $\beta$ is the upper bound for negative similarity.

The cosine similarity $S_{\text{Align}}$ between the text queries and the images is calculated as follows:
\begin{equation}
    S_{\text{Align}}=\cos(\Phi_{\textsc{ENC}}(T),\Phi_{\textsc{ENC}}(I)).
\end{equation}
We define $S_{\text{Align}}^+$ as the similarity between positive image-text pairs, wherein the text and image correspond to the same object. In addition, $S_{\text{Align}}^-$ represents the similarity for negative pairs, where the text and image are related to different objects. 

The alignment loss between original textual and visual features can be denoted as:
\begin{equation}
    \mathcal{L}_{\text{Align}} = \frac{2}{n}\mathcal{L}_{\text{CL}}( S_{\text{Align}}^+, S_{\text{Align}}^-).
\end{equation}

To address the issue of inaccurate representation learning caused by textual ambiguity, we propose a multi-modal reference learning framework consisting of multi-modal reference construction (MMRC) and reference-guided representation learning (RGRL). 
In MMRC, we propose a global fusion module and a local reconstruction module to construct the multi-modal reference with one fusion loss $\mathcal{L}_{\text{Fuse}}$ and another reconstruction loss $\mathcal{L}_{\text{Rec}}$, respectively, which will be introduced exhaustively in Sec.~\ref{latent_ref}.

The constructed multi-modal reference is a comprehensive representation of the target object and can facilitate learning better visual or textual representations. With these references as teachers, we further propose RGRL with a guidance loss $\mathcal{L}_{\text{Guide}}$ to guide the optimization of uni-modal representations, as depicted in Sec.~\ref{reference_learning}. 

In summary, the overall loss of our MMRef can be represented as follows:
\begin{equation}
\label{MMRefloss}
    \mathcal{L} = \mathcal{L}_{\text{Align}}+\lambda_1 \mathcal{L}_{\text{Fuse}}+ \lambda_2 \mathcal{L}_{\text{Rec}} + \lambda_3 \mathcal{L}_{\text{Guide}},
\end{equation}
where $\lambda_1$, $\lambda_2$ and $\lambda_3$ are weights of different losses.

The following parts introduce the detailed implementation of multi-modal reference construction and reference-guided representation learning, as well as the computation of $\mathcal{L}_{\text{Fuse}}, \mathcal{L}_{\text{Rec}}, \mathcal{L}_{\text{Guide}}$, respectively.

\subsection{Multi-Modal Reference Construction}\label{latent_ref}
The multi-modal reference construction consists of a global fusion (GF) module and a local reconstruction (LR) module.
The GF aims to adaptively aggregate a robust multi-modal reference from multiple visual or textual features of the same object using contrastive learning. 
It optimizes the reference at a global level, which may ignore some important local cues of the object. We further utilize the LR to reconstruct the masked textual words conditioned on the multi-modal reference so that the reference can encompass extra local details. \emph{It's worth noting that the multi-modal references are randomly initialized learnable variables.}

\noindent\textbf{Global Fusion:} 
The global fusion module targets aggregating multiple visual and textual features of the same object into a learnable multi-modal reference. 

The multi-modal references $R\in \mathbb{R}^{m\times d}$ are randomly initialized learnable variables, where $m$ is the number of references, equal to the number of objects in the training set. The uni-modal representations $E\in \mathbb{R}^{2n\times d}$ are the combination of visual and textual features, where $E=\Phi_{\textsc{ENC}}(T)\cup\Phi_{\textsc{ENC}}(I)$ and $n$ is the number of image-text pairs in a batch. Besides, $d$ is the dimension of the references or uni-modal representation. In each batch, we sample two image-text pairs for each object.

As shown in Fig.~\ref{Architecture}~(b), to optimize the $i_{th}$ reference $R_i$ as a robust object representation, we draw the reference $R_i$ close to its associated positive uni-modal representations of the same object. This is achieved by maximizing the similarity between them. Besides, to make each reference distinct from the other references, we take the uni-modal representations of other objects as negative samples and minimize the similarity between negative pairs. During this global fusion process, only the learnable references are optimized, while all uni-modal representations are excluded from training through a stop-gradient operation.  

The similarity ${S}_{\text{Fuse}}$ between the multi-modal references and uni-modal representations can be denoted as:
\begin{equation}
    {S}_{\text{Fuse}}=\cos(R, sg(E)),
\end{equation}
where the notation $sg(\cdot)$ is the stop-gradient operation.   

The loss for the global fusion module is as follows:
\begin{equation}
\begin{aligned}
    \mathcal{L}_{\text{Fuse}} = \frac{1}{2n}\mathcal{L}_{\text{CL}}({S}_{\text{Fuse}}^+, {S}_{\text{Fuse}}^-),
\end{aligned}
\label{loss_fuse}
\end{equation}
where $S_{\text{Fuse}}^+$ is the positive similarity between the references and uni-modal representations with the same object label and $S_{\text{Fuse}}^-$ denotes the negative similarity between negative pairs belonging to different object labels.

\noindent\textbf{Local Reconstruction:}
In global fusion, the reference aims to aggregate all uni-modal representations of a target object into itself using contrastive learning. The reference primarily learns salient discriminative details at a global level, which may ignore some important local details. Therefore, a local reconstruction module is designed to incorporate crucial local details into the multi-modal reference. It reconstructs masked words from an obscured textual description, conditioned on the multi-modal reference. If some local details are missed during the global fusion, the reconstruction objective will guide the multi-modal reference to integrate these local details. 

This module is composed of four multi-layer perception (MLP) layers, three attention-based blocks, and a masking language modeling head.
Given an input textual description, we randomly mask out the textual tokens with a ratio of 15\% and replace them with the special token [MASK] following BERT~\cite{devlin-etal-2019-bert}. 
We denote the masked token embeddings of the obscured textual descriptions as $Q$.
The multi-modal references are processed by MLP layers to generate keys ($K$) and values ($V$), which will serve as the reconstruction conditions.

For the attention-based blocks, token embeddings $Q$ first interact with each other to capture the contextual clues in a self-attention layer $\Phi_\textsc{MSA}$ and then pass through the first feed-forward layer $\Phi_\textsc{FFN1}$ to generate contextual queries. The contextual queries are then utilized to search for the overlooked local semantics from $K$ and $V$ through a cross-attention layer $\Phi_\textsc{MCA}$. This process can be denoted as follows:

\begin{equation}
    H=\Phi_\textsc{FFN2}(\Phi_\textsc{MCA}(\Phi_\textsc{FFN1}(\Phi_\textsc{MSA}({Q})), K, V)),
\end{equation}
where $\Phi_\textsc{FFN2}$ is the second feed-forward layer and $H$ is the corresponding output of masked token embeddings.

Finally, a masking language modeling head $\Phi_\textsc{HEAD}$ is utilized to generate the prediction probability $p$ from the output ${O}=\{h_j|h_j\in {H}, j\in {M} \}$ of each masked token and predict the masked words. ${M}$ is the masked position of the textual description. This head comprises a sinle MLP layer. The process can be formulated as follows:
\begin{equation}
    p = \Phi_\textsc{HEAD}({O}).
\end{equation}

The objective of local reconstruction can be represented as the softmax cross-entropy loss, i.e.,
\begin{equation}
    \mathcal{L}_{\text{Rec}} = \text{CrossEntropy}(p, y),
\end{equation}
where $y$ is the index label of the original words in the masked positions.

There is a risk that this module may reintroduce inaccuracies in the text.
The goal of local reconstruction is to guide the reference to focus on important local areas that might be ignored during the global fusion, as shown in Fig.\ref{gf_lr}. Inaccuracy like ``white'' in ``white cap'' shown in Fig.~\ref{introduction} can be depressed by the global fusion, since images provide correct visual cues.

\subsection{Reference-Guided Representation Learning}\label{reference_learning}
Under the guidance of the multi-modal references, the effects of textual ambiguity can be alleviated by aligning uni-modal representations with the references.

To facilitate representation learning with references, we first compute the similarity ${S}_{\text{Guide}}$ between each uni-modal representation and multi-modal reference. Different from the process of global fusion, the multi-modal references are detached from training by a stop-gradient operation while the uni-modal representations are learnable in RGRL.
We could get the positive similarity ${S}_{\text{Guide}}^+$ between each uni-modal representation and its corresponding multi-modal reference belonging to the same object, and the negative similarity ${S}_{\text{Guide}}^-$ with its negative multi-modal references. 

The optimization goal of RGRL is to maximize the positive similarity ${S}_{\text{Guide}}^+$ to provide positive guidance. Meanwhile, it reduces the negative similarity, ${S}_{\text{Guide}}^-$, ensuring that uni-modal representations are clearly distinguished from the references of other objects. 

To maintain the uni-modal representations within the existing unified feature space, we use the same contrastive loss as defined in Eq.~(\ref{loss_func}), rather than other contrastive losses such as infoNCE~\cite{infoNCE} or margin ranking loss~\cite{marginloss}. As shown in Tab.~\ref{exp_guide_loss}, using other contrastive losses may harm representation learning by pushing uni-modal representations into other feature spaces.
The loss $\mathcal{L}_{\text{Guide}}$ is denoted as follows:
\begin{equation}
    \mathcal{L}_{\text{Guide}} = \frac{1}{2n}\mathcal{L}_{\text{CL}}({S}_{\text{Guide}}^+,{S}_{\text{Guide}}^-).
\label{loss_guide}
\end{equation}

\subsection{Inference with Reference-based Refinement}\label{inference}
Each multi-modal reference has contained comprehensive semantics of a specific object. Besides guiding the optimization of uni-modal representation in the training phase, these multi-modal references can also be utilized during the inference stage to refine retrieval results. Multi-modal references can be regarded as shared semantic prototypes that bridge the visual and textual modalities. By leveraging these references as an intermediary, we can map the original visual or textual features into a unified multi-modal space and compute the reference-based similarity between them. This unified multi-modal space could effectively narrow the gap between visual and textual modalities.

\begin{figure}[t]
    \begin{center}
        \includegraphics[width=0.95\linewidth]{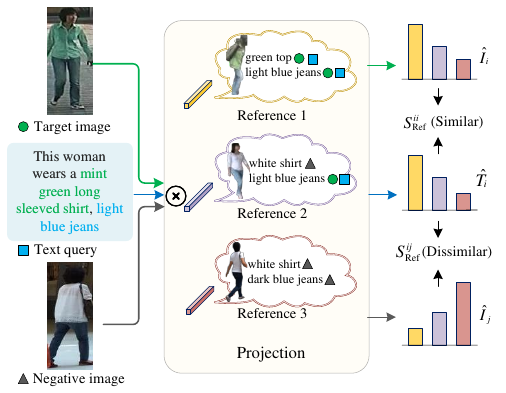}
    \end{center}
    \caption{Illustration of reference-based similarity in reference space. Textual or visual features are projected into a shared reference space, where modality-agnostic semantics are preserved and modality-specific noises are discarded. The reference-based similarity is utilized to refine the initial similarity.}\label{reference_refine}
\end{figure}

As illustrated in Fig.~\ref{reference_refine}, given a text query, if its textual feature closely aligns with certain multi-modal references, the visual feature of its depicted object will also align with these references.
We hence could use those multi-modal references as prototypes, and project the original textual features $\Phi_{\textsc{ENC}}(T)$ and visual features $\Phi_{\textsc{ENC}}(I)$ into a unified multi-modal reference space, leading to reference-based textual features $\hat{T}\in\mathbb{R}^{{n}\times m}$ and visual features $\hat{I}\in\mathbb{R}^{{n}\times m}$, i.e.,
\begin{equation}
\label{projection}
    \hat{I} = \Phi_{\textsc{ENC}}(I) R^{\top},   \ \ \ \  \hat{T} = \Phi_{\textsc{ENC}}(T) R^{\top}.
\end{equation}

This projection preserves the modality-agnostic semantics while discarding modality-specific details, such as visual backgrounds and textual function words. Therefore, we compute the reference-based similarity $S_{\text{Ref}}$ between projected textual and visual features to refine the original similarity, which can be computed as:
\begin{equation}
\label{ref_similarity}
    S_{\text{Ref}} = \cos(\hat{T},\hat{I}).
\end{equation}

During inference, the reference-based similarity is merged with the initial similarity $S_{\text{Algin}}$ in the original space to get the final results of retrieval. The refined similarity for text-to-image retrieval can be denoted as follows:
\begin{equation}
    S_{\text{Final}}=S_{\text{Align}}+wS_{\text{Ref}},
\end{equation}
where $w$=0.5 is a fusion weight of reference-based similarity.

\begin{figure*}
	\begin{center}
		\includegraphics[width=0.95\linewidth]{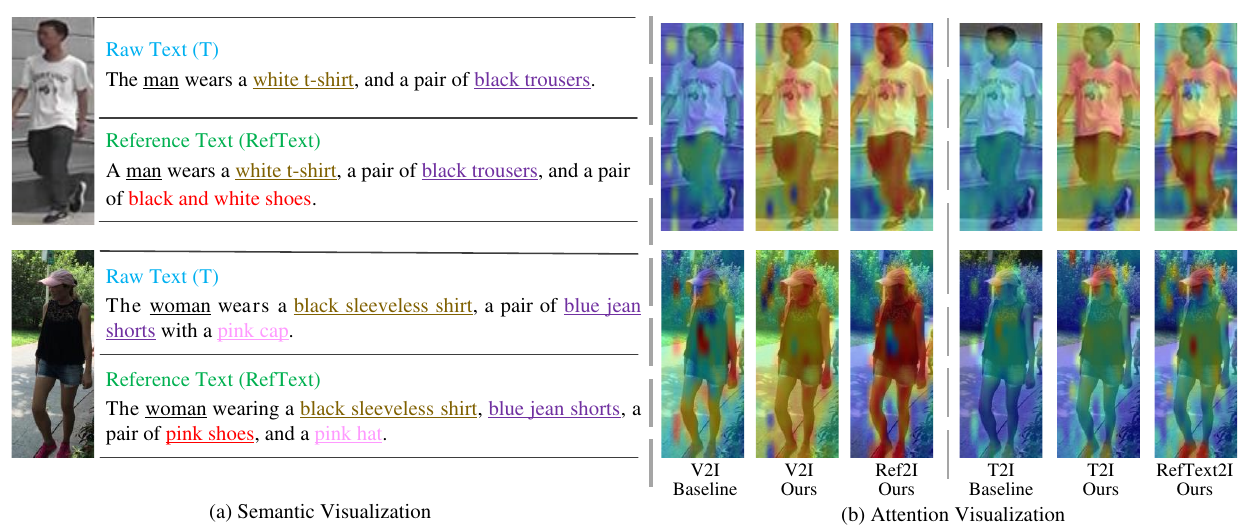}
	\end{center}
 \vspace{-5pt}
	\caption{
		 Visualization of multi-modal reference. (a) “Raw Text” denotes the original caption. “Reference Text” is a semantic caption of the multi-modal reference, which is more complete and accurate. Textual phrases in red color are descriptions that do not appear in the raw text. (b) For each identity, “V2I” illustrates the attention of visual representation on the given image. “Ref2I” denotes the attention of reference embedding on the image. “T2I” is the text-to-image attention of raw text, and “RefText2I” is the attention of feature extracted from the reference text generated by captioning the reference embedding. Both “Ref2I” and “RefText2I” demonstrate that our reference encompasses more meaningful details of the person.}
	\label{gpt_caption}
\end{figure*}

\subsection{Visualization of Multi-modal Reference} 
The multi-modal reference is expected to encompass all essential details of a target object and guide learning better uni-modal representations.
Fig.~\ref{gpt_caption} verify it with semantic visualization and attention visualization, respectively. 

\noindent\textbf{Semantic Visualization:} In order to show semantics for the multi-modal reference, we fine-tune an image captioning model ClipCap~\cite{mokady2021clipcap} on CUHK-PEDES, whose key idea is to take the frozen visual feature extracted by our MMRef as a prefix embedding to prompt GPT2 ~\cite{radford2019language} to generate a corresponding caption. 
Since the multi-modal reference embedding is aligned with the uni-modal visual representations in our framework, we can leverage this fine-tuned ClipCap to generate captions for any multi-modal reference. As shown in Fig.~\ref{gpt_caption}~(a), the text of multi-modal reference~(RefText) is more complete and accurate than the raw text of given images. We thus conclude that the multi-modal reference is a comprehensive and accurate representation of the target object. 

\noindent\textbf{Attention Visualization:} We further visualize the attention of baseline visual representation, MMRef visual representation, and multi-modal reference in the first three columns of Fig.~\ref{gpt_caption}~(b). As shown in the column ``Ref2I'', the multi-modal reference embedding accurately pays attention to salient parts of the person. Comparing ``V2I'' between the baseline and ours, we find that the visual representation of our method can depict more discriminative areas and details of the person. It means that the learned reference embedding is meaningful and can guide the learning of better visual representations. 

We further visualize the text-to-image cross-modal attention in the last three columns of Fig.~\ref{gpt_caption}~(b). It shows that our method pays attention to most of the areas mentioned in the raw text or reference text. For example, the feature of reference text pays more attention to the ``shoes'' area, which is not mentioned in the raw text. The ``T2I'' comparison between the baseline and ours shows that the textual representations of our MMRef depicts more discriminative regions, indicating that the reference leads to better textual representations and cross-modality alignment. The comparison between ``Ref2I'' and ``RefText2I'' indicates that areas of ``shirt'' and ``jean shorts'' in ``RefText2I'' have lower attention weights. It's reasonable because ``black shirt‘’ and ``jean shorts'' can be commonly observed across different persons, and thus are less discriminative attributes than ``pink shoes'' and ``pink hat''. 

These visualizations verify the effectiveness of our proposed multi-modal reference construction and reference-guided representation learning. More extensive experiments will be conducted in the following section.

\section{Experiments}
This section further validates the effectiveness of our proposed MMRef. First, we introduce the experimental setup including datasets, evaluation metrics, and implementation details. Then we compare our method with recent works to show the effectiveness of our framework. Next, we conduct ablation studies to validate the effectiveness of each module. Furthermore, we illustrate the multi-modal reference in both visual and textual ways and explain the improvement of our method. Finally, we demonstrate that aligning visual features with text descriptions can help improve the domain generalization capability for image-based person retrieval.

\begin{table*}[t]
\caption{Comparison with the state-of-the-art methods on CUHK-PEDES~\cite{li2017person}. R@1, R@5, and R@10 for text-to-image and image-to-text tasks are reported. The best results are bold.} \label{tabel1}
\centering
\small
\setlength{\tabcolsep}{12pt}
\begin{tabular}{c|c|c|ccc|ccc}
\hline
 \multirow{2}{*}{Methods} & \multirow{2}{*}{Source} & \multirow{2}{*}{\makecell{Visual\\Backbone}} & \multicolumn{3}{c|}{Text-to-Image} & \multicolumn{3}{c}{Image-to-Text} \\ \cline{4-9}
 & & & R@1 & R@5 & R@10 & R@1 & R@5 & R@10 \\ \hline\hline
GNA-RNN~\cite{li2017person} & CVPR'17 & VGG16 & 19.05 & - & 53.64 &  \\ 
 GLA~\cite{chen2018improving} & ECCV'18 & ResNet50 & 43.58 & 66.93 & 76.26 &  \\ 
 Dual Path~\cite{zheng2020dual}& TOMM'20 & ResNet50 & 44.40 & 66.26 & 75.07 &  \\
 CMPM/C~\cite{zhang2018deep} & ECCV'18 & MobileNet & 49.37 & 71.69 & 79.27 & 60.96 & 84.42 & 90.83 \\
 AATE~\cite{AATE} & TMM'20 & ResNet50& 52.42 & 74.98 & 82.74 & \\ 
 MIA~\cite{niu2020improving} & TIP'20 & ResNet50 & 53.10 & 75.00 & 82.90 &  \\
PMA~\cite{jing2020pma} & AAAI'20 & ResNet50 & 53.81 & 73.54 & 81.23 &  \\
TIMAM~\cite{sarafianos2019adversarial} & ICCV'19 & ResNet101 & 54.51 & 77.56 & 84.78 & 67.40 & 88.65 & 93.91 \\
ViTAA~\cite{wang2020vitaa} & ECCV'20 & ResNet50 & 55.97 & 75.84 & 83.52 & 65.71 & 88.68 & 93.75 \\
DSSL~\cite{zhu2021dssl} & MM'21 & ResNet50 & 59.98 & 80.41 & 87.56 &  \\
SSAN~\cite{ding2021semantically} &arXiv'21 & ResNet50 & 61.37 & 80.15 & 86.73 & \\
TextReID~\cite{han2021textreid} & BMVC'21 & ResNet50 & 61.65 & 80.98 & 86.78 & 75.96 & 93.40 & 96.55\\
ACSA~\cite{ji2022asymmetric} & TMM'23 & Swin-Tiny & 63.56 & 81.40 & 87.70 & \\
LBUL~\cite{wang2022look} &MM'22 & ResNet50 & 64.04 & 82.66 & 87.22 &  \\
LGUR~\cite{shao2022learning} & MM'22 & ResNet50 & 64.21 & 81.94 & 87.93 &  \\
TIPCB ~\cite{chen2022tipcb} & Neuro'22& ResNet50 & 64.26 & 83.19 & 89.10 & 73.55 & 92.26 & 96.03 \\
CAIBC~\cite{wang2022caibc} & MM'22 & ResNet50 & 64.43 & 82.87 & 88.37 &\\
PBSL~\cite{PBSL} & MM'23 & ResNet50 & 65.32 & 83.81 & 89.26 & \\
MMRef(Ours) & This Paper & RetNet50 & \textbf{66.15} & \textbf{84.73} & \textbf{90.29} & \textbf{80.71}& \textbf{95.58} & \textbf{97.76} \\ \hline
IVT~\cite{shu2023see} & ECCVW'22 & ViT-B & 65.59 & 83.11 & 89.21 &  \\ 
TransTPS~\cite{TransTPS} & TMM'24 & ViT-B & 68.23 &  86.37 & 91.65 & \\
MMGCN~\cite{MMGCN} & TMM'24 & GNN & 69.40 & 87.07 & 90.82 &  \\ 
CFine~\cite{yan2022cfine} & TIP'23 & ViT-B& 69.57 & 85.93 & 91.15 &  \\ 
VGSG~\cite{he2023vgsg} & TIP'23 & ViT-B & 71.38& 86.75 & 91.86 & 84.92 & 96.35 & 98.24 \\

MMRef(Ours) & This Paper & ViT-B & \textbf{72.25} & \textbf{88.24} & \textbf{92.61} & \textbf{85.98} & \textbf{97.01} & \textbf{98.93}\\ \hline
\end{tabular}
\end{table*}

\subsection{Experimental Setup}
\textbf{Datasets for Text-to-Person Retrieval:} 
We first evaluate our approach on three text-to-person retrieval datasets: CUHK-PEDES, ICFG-PEDES, and RSTPReid. CUHK-PEDES~\cite{li2017person} includes 40,206 images and 80,440 text descriptions of 13,003 persons. It is split into 11,003 training identities with 68,126 image-text pairs, 1,000 validation persons with 6,158 image-text pairs, and 1,000 test individuals with 6,156 image-text pairs. ICFG-PEDES~\cite{ding2021semantically} is a new database that contains 54,522 text descriptions for 54,522 images of 4,102 persons collected from the MSMT17~\cite{wei2018person} dataset. It is split into a training set with 34,674 image-text pairs of 3,102 persons, and a testing set with 19,848 image-text pairs for the remaining 1,000 persons. RSTPReid~\cite{zhu2021dssl} is also constructed based on MSMT17~\cite{wei2018person}, which includes 41,010 text descriptions and 20,505 images of 4,101 persons. Each person contains 5 images caught by 15 cameras and each image corresponds to 2 text descriptions. The training, validation, and testing sets have 3,701, 200, and 200 identities, respectively.

\textbf{Datasets for Text-to-Bird Retrieval:} 
The Caltech-UCSD Birds (CUB)~\cite{WahCUB_200_2011,reed2016learning} dataset consists of 11,788 bird images from 200 different categories. Each image is labeled with 10 visual descriptions. The dataset is split into 100 training, 50 validation, and 50 test categories. 

\textbf{Datasets for Text-to-Flower Retrieval:}
The Oxford102 Flowers (Flowers)~\cite{nilsback2008automated,reed2016learning} dataset contains 8,189 flower images of 102 different categories, and each image has 10 textual descriptions. The data splits provide 62 training, 20 validation, and 20 test categories.

\textbf{Datasets for Image-based Person Retrieval:} To validate our MMRef can help improve the model's domain generalization capability in image-based person retrieval, we also conduct experiments on commonly used datasets Market1501~\cite{zheng2015scalable} and MSMT17~\cite{wei2018person}. Market1501 is a large-scale dataset captured from 6 cameras, containing 32,668 images with 1,501 identities. It is divided into 12,936 images of 751 identities for training and 19,732 images of 750 identities for testing.
MSMT17 is another widely used person Retrieval dataset. It contains 126,441 images of 4,101 identities captured from 15 cameras. It is divided into 32,621 images of 1,041 identities for training and 93,820 images of 3,060 identities for testing.

\textbf{Evaluation Metrics:} To ensure a fair comparison with the previous methods, we report R@K(K=1,5,10)~\cite{karpathy2015deep} when compared with state-of-the-art models following previous works~\cite{wang2022caibc,niu2020improving}. It reports the percentage of the images where at least one corresponding concept is retrieved correctly among the top-K results. Additionally, we report the mean average precision (mAP), the average precision across all queries, in ablation studies for analysis and future comparison. In CUB and Flowers experiments, we report AP@50 following~\cite{reed2016learning, li2017identity, zhang2018deep}, which represents the percent of top-50 scoring images whose class matches that of the text query, averaged over all the test classes.

\begin{table}[t]
\caption{Comparison with the state-of-the-art methods on text-to-image task of ICFG-PEDES ~\cite{ding2021semantically}. }\label{tabel2}
\centering
\small
\setlength{\tabcolsep}{14pt}
\begin{tabular}{c|ccc}
\hline
Methods  & R@1 & R@5 & R@10\\ \hline\hline
Dual Path~\cite{zheng2020dual} & 38.99 & 59.44  & 68.41\\ 
MIA~\cite{niu2020improving} & 46.49 & 67.14 & 75.18 \\
SCAN~\cite{lee2018stacked} & 50.05 & 69.65 & 77.21 \\
ViTAA~\cite{wang2020vitaa} & 50.98 & 68.79 & 75.78 \\
SSAN~\cite{ding2021semantically}  & 54.23 & 72.63 & 79.53 \\
TIPCB~\cite{chen2022tipcb}  & 54.96 & 74.72 & 81.89 \\
IVT~\cite{shu2023see}  & 56.04 & 73.60 & 80.22 \\
SRCF~\cite{suo2022simple} & 57.18 & 75.01& 81.49 \\
CFine~\cite{shu2023see}  & 60.83 & 76.55 & 82.42 \\ \hline
MMRef (Ours) & \textbf{63.50} & \textbf{78.19} & \textbf{83.73} \\ \hline
\end{tabular}
\end{table}

\begin{table}[t]
\caption{Comparison with state-of-the-art methods on text-to-image task of RSTPReid ~\cite{zhu2021dssl}.}\label{tabel3}
\centering
\small
\setlength{\tabcolsep}{15pt}
\begin{tabular}{c|ccc}
\hline
Methods  & R@1 & R@5 & R@10\\ \hline\hline
 AMEN~\cite{wang2021amen} & 38.45 & 62.40 & 73.80\\ 
 DSSL~\cite{zhu2021dssl} & 39.05 & 62.60 & 73.95 \\
 SUM~\cite{wang2022sum} & 41.38 & 67.48 & 76.48 \\
 SSAN~\cite{ding2021semantically} & 43.50 & 67.80 & 77.15 \\
 LBUL~\cite{wang2022look} & 45.55 & 68.20 & 77.85 \\
 IVT~\cite{shu2023see} & 46.70 & 70.00 & 78.80 \\
 CFine~\cite{yan2022cfine}  & 50.55 & 72.50 & 81.60 \\\hline
{MMRef (Ours)} & \textbf{56.20} & \textbf{77.10} & \textbf{85.80} \\ \hline
\end{tabular}
\end{table}

\textbf{Implementation Details:}
We conduct experiments on two NVIDIA 3090 GPUs based on PyTorch.
To ensure a 
fair comparison with existing approaches, we adopt ResNet50~\cite{he2016deep} and ViT-B/16~\cite{dosovitskiy2021vit} from CLIP~\cite{clip_radford2021learning} as the visual backbone. For ResNet50, the input resolution of the image is 384$\times$128 and the dimension $D$ of visual features is 1024. For ViT-B/16, all images are resized to 224$\times$224 and the dimension $D$ is 512.
In addition, we employ random horizontal flipping as image augmentation. The text encoder is initialized with the transformer in~\cite{clip_radford2021learning} and the input length of textual token sequences is 77. Following~\cite{wang2020vitaa, han2021textreid}, the hyperparameters in the loss function are set as: $\tau_p$ = 10 and $\tau_n$ = 40. The $\alpha$ and $\beta$ for different datasets are introduced in Fig.~\ref{exp_alpha_ablation}. The losses weights $\lambda_1$, $\lambda_2$, and $\lambda_3$ are set to 0.25, 0.25, and 4, repectively.
Our proposed MMRef model is trained in an end-to-end manner for 20 epochs.
The parameters are optimized by Adam~\cite{adam2014} with 2 warm-up epochs and linear learning rate decay. For each identity, two image-text pairs are sampled in one iteration. A batch consists of 90 image-text pairs belonging to 45 different identities. The peak learning rate is set to $4e^{-5}$.
MMRef supports multiple hardware platforms and already supports training and deployment on the Ascend 910B NPU.

\subsection{Comparison with State-of-the-art Methods}
In this section, we compare the performance of our proposed MMRef with state-of-the-art methods on CUHK-PEDES~\cite{li2017person}, ICFG-PEDES~\cite{ding2021semantically}, RSTPReid~\cite{zhu2021dssl} in the person retrieval, CUB~\cite{WahCUB_200_2011} in bird retrieval, and Flowers~\cite{nilsback2008automated} in flower retrieval. 
Our MMRef consistently achieves promising performance on those datasets.

\begin{table}[t]
\setlength{\tabcolsep}{2pt}
\centering
\caption{Comparison of text-to-image retrieval AP@50 on the CUB and Flowers dataset and image-to-text R@1 on CUB.}
\small
\label{cub_and_flower}
\begin{tabular}{c|c|cc|c}
\hline
\multirow{2}{*}{Method} & \multirow{2}{*}{Backbone} & \multicolumn{2}{c|}{Text-to-Image} & \multicolumn{1}{c}{Image-to-Text} \\ 

&   & \multicolumn{1}{c}{CUB} & \multicolumn{1}{c|}{Flowers} & CUB\\
\hline \hline
Word2Vec~\cite{mikolov2013distributed}    &  -  & 33.5 & 52.1& 38.6 \\
Word CNN~\cite{reed2016learning}     &  GoogleNet & 43.3  & 56.3 & 51.0 \\
Word CNN-RNN~\cite{reed2016learning}  & GoogleNet  & 48.7 & 59.6 & 56.8 \\
GMM+HGLMM~\cite{klein2015associating}   & GoogleNet & 35.6  & 52.8 & 36.5\\
Triplet~\cite{li2017identity}     & GoogleNet & 52.4  & 64.9& 52.5 \\
Co-attention~\cite{li2017identity} & GoogleNet & 57.6 & 70.1 & 61.5 \\
CMPM/C~\cite{zhang2018deep}  & ResNet50  & 67.9 & 69.7 & 64.3 \\
AATE~\cite{AATE} & ResNet50 & 71.5 & - & 65.8 \\
\hline
MMRef(Ours)   & ResNet50  & \textbf{72.4} & \textbf{76.5} &\textbf{66.3} \\ 
MMRef(Ours)   & ViT-B & \textbf{87.0} & \textbf{83.6} & \textbf{68.7} \\ \hline
\end{tabular}
\end{table}
\textbf{Person Retrieval:} 
We first evaluate our MMRef on CUHK-PEDES. 
As is shown in Tab.~\ref{tabel1}, our MMRef has achieved promising performance with either ResNet50 or ViT-B/16 backbone. Specifically, among ResNet50-based methods, our MMRef outperforms the recent method PBSL~\cite{PBSL} which uses an additional graph neural network to compute extra region-word similarity. In addition, our MMRef(ResNet50) has only 103M parameters which is much smaller than other recent ResNet50-based methods. For instance, CABIC and TIPCB have 160M and 185M parameters, respectively.
Compared with the recent ViT-based works CFine and VGSG that also utilize the CLIP to initialize backbones, our MMRef consistently achieves better performances on all metrics, demonstrating the effectiveness and superiority of our method.

To further validate our proposed method, we also compare ViT-based MMRef against the previous works on two other benchmarks in person retrieval. As shown in Tab.~\ref{tabel2}, our MMRef consistently achieves better performance than other methods in all evaluation metrics on ICFG-PEDES. Besides, as shown in Tab.~\ref{tabel3}, MMRef outperforms recent CFine by a large margin on the RSTPReid dataset and obtains 56.20\%(+5.65\%), 77.10\%(+4.60\%) and 85.80\%(+4.20\%) of R@1, R@5 and R@10 accuracy. Results on these benchmarks demonstrate the effectiveness and robustness of MMRef.

\textbf{Other Fine-grained Retrieval:} We further validate our MMRef in the text-to-bird retrieval on the CUB dataset and text-to-flower retrieval on the Flowers dataset. As illustrated in Tab.~\ref{cub_and_flower}, Our ResNet50-based MMRef outperforms the recent AATE on the CUB dataset, and also surpasses the CMPM/C on the Flowers dataset by 6.8\% in AP@50. In addition, the ViT-based MMRef achieves 87.0\% and 83.6\% AP@50 on CUB and Flowers, respectively, which boosts the state-of-the-art performance on those datasets by a notable margin. 

\textbf{Image-to-Text Retrieval:}
Our MMRef can be directly extended to perform image-to-text retrieval. In Tab.~\ref{tabel1}, our MMRef achieves superior image-to-text performance in both ResNet50 and ViT-B comparisons.
Besides, as shown in Tab.~\ref{cub_and_flower}, we also achieve better image-to-text R@1 accuracy in the CUB dataset.

\subsection{Ablation Study of Components}
This part evaluates the effectiveness of our method by gradually adding the global fusion module, local reconstruction module, and reference-based refinement into the baseline and showing the performance improvement. It's worth noting that reference-guided representation learning (RGRL) is an essential component of the multi-modal reference learning framework. Either the global fusion or the local reconstruction relies on the RGRL to demonstrate its effectiveness.

\begin{table}[t]
\caption{Ablation study on each component of MMRef. ``RGRL'' denotes uni-modal representation learning with reference guiding, an essential module of MMRef. ``GF'' means the global fusion module. ``LR'' represents the local reconstruction module. ``REFINE'' represents the reference-based refinement method. The best results are highlighted in bold, whereas the second-best results are underscored for emphasis. } \label{tabel4}
\small
\centering
\setlength{\tabcolsep}{3pt}
\begin{tabular}{c|cccc|cccc}
\hline
\multirow{2}{*}{Settings} &\multirow{2}{*}{RGRL} &\multirow{2}{*}{GF} & \multirow{2}{*}{LR} & \multirow{2}{*}{REFINE} & \multicolumn{4}{c}{CUHK-PEDES} \\
& & & & & R@1 & R@5 & R@10 & mAP \\ \hline\hline
 {Baseline} & & & & & 68.47 & 85.28 & 90.76 & 60.41 \\ 

 A & \checkmark &\checkmark & & & 70.84 & 87.05 & 92.04 & 64.21 \\
  {B} & \checkmark &\checkmark & & \checkmark & 71.44 & 87.00 & 92.14 & 64.60 \\ 
{C} & \checkmark &\checkmark & \checkmark&  & \underline{71.73} & \underline{87.80} & \underline{92.37} & \underline{64.61}  \\ 
  MMRef& \checkmark &\checkmark & \checkmark& \checkmark & \textbf{72.25} & \textbf{88.24} & \textbf{92.61} & \textbf{65.23} \\ 
  \hline
\end{tabular}
\end{table}

\begin{figure}
	\begin{center}
		\includegraphics[width=0.75\linewidth]{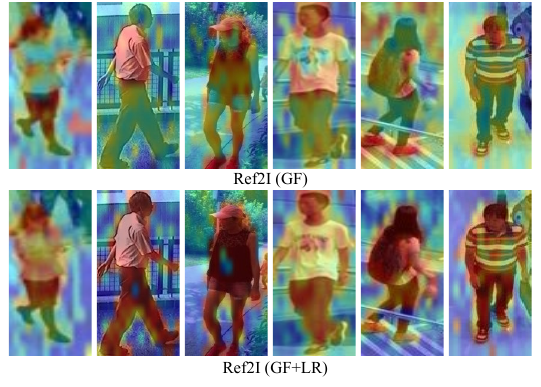}
	\end{center}
	\caption{
		{Attention visualization of references constructed using global fusion (GF) or a combination of global fusion and local reconstruction (GF+LR).} }
	\label{gf_lr}
\end{figure}

\begin{table}[t]
    \caption{Analysis of hyperparameter $w$ on CUHK-PEDES. }\label{hyperparameter table}
   \small
   \centering
   \setlength{\tabcolsep}{6pt}
   \begin{tabular}{c|cccc}
   \hline
   $w$ & R@1 & R@5  & R@10 & mAP\\ \hline\hline
      0 & 71.73 & 87.80 & 92.37 & 64.61 \\ \hline
      0.1 & 71.86 & 88.04 & 92.51 &  64.86\\ 
      0.3 & {72.24} & \textbf{88.27} & \textbf{92.67} & 65.18 \\
      0.5 & \textbf{72.25} & {88.24} & {92.61} & \textbf{65.23} \\
      0.7 & 72.04 & 88.17 & 92.58 & {65.20} \\ 
      0.9 & 71.85 & 87.88 & 92.51 & 65.12 \\ \hline
   \end{tabular}
\end{table}

\begin{table}[t]
\centering
\caption{Inference time with reference-based refinement versus its absence. ``total'' denotes the total inference time of all text queries. ``per-query'' represents the inference time of a single text query. }
\small
\label{computational_burden}
\begin{tabular}{c|cc|cc}
\hline
\multirow{2}{*}{} & \multicolumn{2}{c|}{CUHK-PEDES} & \multicolumn{2}{c}{ICFG-PEDES} \\ \cline{2-5} 
                  & total           & per-query          & total          & per-query          \\ \hline
w/o REFINE  &   26.38s           &  0.0043s               &     120.94s        &     0.0061s           \\ \hline
MMRef             &   26.79s           &  0.0044s               &     122.33s         &     0.0062s            \\ \hline
\end{tabular}

\end{table}

\textbf{Baseline:} We initialize the text encoder and image encoder with the backbones from CLIP~\cite{clip_radford2021learning} and construct a baseline by fine-tuning a few epochs with an aligning loss $\mathcal{L}_{\text{Align}}$. 

\textbf{Effectiveness of Global Fusion:}
The global fusion module aims to construct a comprehensive multi-modal representation by globally aggregating all uni-modal representations with the fusing loss $\mathcal{L}_{\text{Fuse}}$. The constructed reference then guides learning better uni-modal representations with RGRL, which is an essential part of our framework. Comparing ``Baseline'' and ``A'' in Tab.~\ref{tabel4}, the global fusion module boosts the performances on all retrieval metrics. The ``Ref2I(GF)'' in Fig.~\ref{gf_lr} shows that the reference pays attention to most of the discriminative parts of the target object.

\textbf{Effectiveness of Local Reconstruction:}
The local reconstruction module utilizes the reference as a condition to predict masked textual words. It refines the multi-modal reference to encompass more local details. As demonstrated by the comparison between ``C'' and ``A'' in Table~\ref{tabel4}, all evaluation metrics have shown a notable improvement. As shown in Fig.~\ref{gf_lr}, the reference encompasses more details and depicts most of the body foreground of the person after introducing the local reconstruction module. The experiment results and visualization demonstrate the effectiveness of the local reconstruction module.

\textbf{Effectiveness of Reference-based Refinement:} 
The multi-modal references can also be utilized as multi-modal semantic prototypes to refine the initial results during inference.
By comparing ``B'' with ``A'', or ``MMRef`` with ``C'', we conclude that the reference-based refinement method consistently improves performance across all metrics. This indicates that multi-modal references not only help in learning better uni-modal representations during training, but also facilitate the inference process. 

In Table~\ref{hyperparameter table}, we conduct ablation experiments on the hyperparameter $w$ on CUHK-PEDES. We set $w$ to 0.5 to achieve the best R@1 and mAP performance. The additional computation overheads of reference-based refinement are simply the projection operation in Eq.~(\ref{projection}) and the dot product operation in Eq.~(\ref{ref_similarity}). As shown in Tab.~\ref{computational_burden}, the additional inference time for each text query is $1e^{-4}$. The additional total inference time of all text queries is also marginal. It demonstrates that the additional computational cost of reference-based refinement could be considered negligible.

\begin{figure}
	\begin{center}
		\includegraphics[width=1\linewidth]{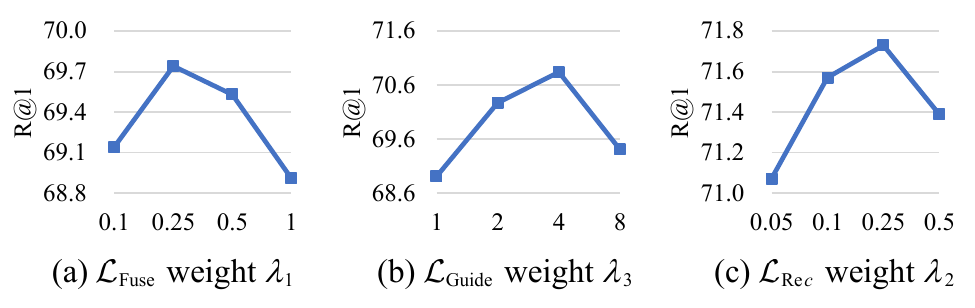}
	\end{center}
	\caption{{Ablation studies of loss weights, $\lambda_1$ of global fusion loss $\mathcal{L}_\text{Fuse}$, $\lambda_3$ of RGRL loss $\mathcal{L}_\text{Guide}$, and $\lambda_2$ of local reconstruction loss $\mathcal{L}_\text{Rec}$. Experiments are conducted on the CUHK-PEDES dataset without reference-based refinement.}}
	\label{loss_weights}
\end{figure}

\begin{table}[t]
\caption{Ablation Study of Different RGRL Losses on CUHK-PDDES without Reference-based Refinement Module}.\label{exp_guide_loss}
\centering
\small
\begin{tabular}{c|ccc}
\hline
RGRL Loss  & R@1 & R@5 & R@10\\ \hline\hline
 infoNCE~\cite{infoNCE} & 68.84 & 86.22 & 91.16 \\
 Margin Ranking Loss~\cite{marginloss} & 69.95 & 86.84 & 91.46 \\\hline
{ $\mathcal{L}_{\text{Guide}}$ (Ours)} & \textbf{71.73} & \textbf{87.80} & \textbf{92.37} \\ \hline
\end{tabular}
\end{table}

\subsection{Ablation Study of Loss Weights} \label{Loss_ablation_sec}

This part further studies loss weights in Eq.~(\ref{MMRefloss}). 
The weight of the fundamental alignment loss $\mathcal{L}_\text{Algin}$ is set to 1.0. Ablation studies are conducted by modifying the other three weights. We illustrate the experimental results in Fig.~\ref{loss_weights}.

With $\lambda_2$=0 and $\lambda_3$=1, we first test the weight $\lambda_1
$ of the global fusion loss $\mathcal{L}_\text{Fuse}$. $\lambda_1$ controls the optimization speed of the reference. As shown in Fig.~\ref{loss_weights}~(a), our method achieves the best R@1 accuracy with $\lambda_1$=0.25. It demonstrates that the multi-modal reference aggregates details of the target object from multiple batches and optimization steps. A too small $\lambda_1$, e.g., 0.1, may lead to a slow optimization to the reference, and is not effective for learning the uni-modal representations.

We proceed to test the loss weight $\lambda_3$ of RGRL loss $\mathcal{L}_\text{Guide}$ in Fig.~\ref{loss_weights}~(b), by fixing $\lambda_1$=0.25 and $\lambda_2$=0. $\lambda_3$  determines the strength of guidance in reference-guided representation learning (RGRL). Our method performs best with $\lambda_3$=4. This large weight $\lambda_3$ provides strong guidance during training, and validates the effectiveness of RGRL. Too large $\lambda_3$ degrades the performance because the randomly initialized reference may provide wrong guidance at the first several epochs.

With $\lambda_1$=0.25 and $\lambda_3$=4, we finally study the loss weight $\lambda_2$ of local reconstruction loss $\mathcal{L}_\text{Rec}$ in Fig.~\ref{loss_weights}~(c). The local reconstruction module performs best with $\lambda_2$=0.25. Setting $\lambda_2$ to other values still consistently outperforms the setting of fixing it to 0. The global fusion and local reconstruction are in balance when both $\lambda_1$ and $\lambda_2$ are equal to 0.25.

\subsection{Ablation Study of Smaller Modules}
We ablate the smaller modules of MMRef in this subsection. It's worth noting that the following ablation experiments are conducted on the CUHK-PEDES dataset without the reference-based refinement module.

\textbf{Loss Selection of RGRL:}
In reference-guided representation learning, we utilize the contrastive loss introduced in Eq.~(\ref{loss_guide}). In this equation, the references are detached from the training process, allowing only the uni-modal representations to be optimized under the guidance of these references. Differently, Eq.~(\ref{loss_fuse}) is designed to construct a robust reference for each target object. Therefore in Eq.~(\ref{loss_fuse}), the references are learnable, and the uni-modal representations are detached by stopping the gradient. Although both Eq.~(\ref{loss_fuse}) and Eq.~(\ref{loss_guide}) are computed based on contrastive loss, they present different optimization parameters and goals. In Tab.~\ref{exp_guide_loss}, we conduct experiments to compare $\mathcal{L}_{\text{Guide}}$ with other losses, i.e., infoNCE and margin ranking loss. The temperature in infoNCE is set to 100.0 and the margin of margin ranking loss is 0.2. The results demonstrate our $\mathcal{L}_{\text{Guide}}$ performs the best. This could be because the loss function and hyper-parameters of $\mathcal{L}_{\text{Guide}}$ are kept the same as alignment loss $\mathcal{L}_{\text{Align}}$ and global fusion loss $\mathcal{L}_{\text{Fuse}}$, ensuring the uni-modal representations are optimized in the existing unified feature space. Utilizing other contrastive losses may alter the existing feature space, and degrade the performance. 

\begin{figure}[t]
    \begin{center}
        \includegraphics[width=0.9\linewidth]{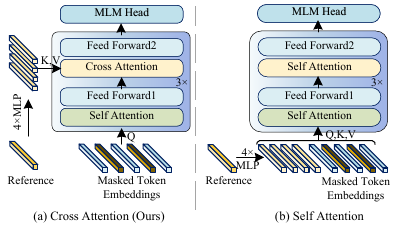}
    \end{center}
    \caption{Two different architectures of the local reconstruction module.}\label{exp_mlm_design}
\end{figure}

\begin{figure}[t]
	\begin{center}
		\includegraphics[width=1\linewidth]{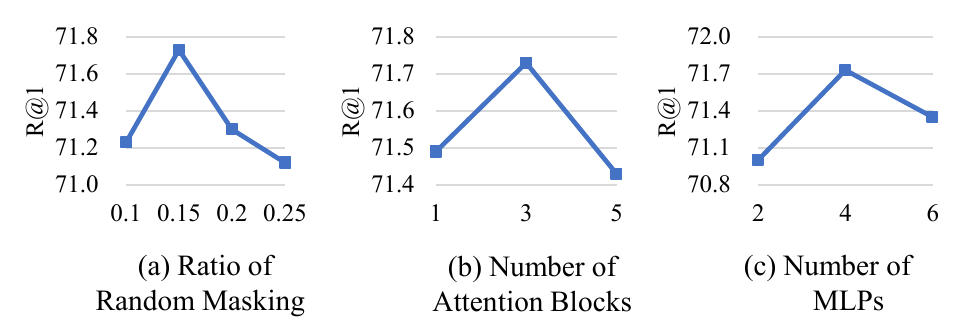}
	\end{center}
	\caption{{Ablation studies of hyper-parameters in the local reconstruction on the CUHK-PEDES without reference-based refinement, including the ratio of random masking, the number of attention blocks, and the number of MLPs.}}
	\label{exp_mlm_ablation}
\end{figure}

\textbf{Architecture Design of LR:}
As shown in Fig.~\ref{exp_mlm_design}, we implement two different architectures for the local reconstruction module. Our cross-attention version in Fig.~\ref{exp_mlm_design}~(a) utilizes the projected reference features as {K}ey-{V}alue and the masked token embeddings as {Q}uery. This method achieves 71.73\% R@1 accuracy. The self-attention version in Fig.~\ref{exp_mlm_design}~(b) concatenates the projected reference features and masked token embeddings as input and leverages them as {Q}uery-{K}ey-{V}alue. It requires a larger computation budget, and achieves a lower R@1 accuracy of 71.25\%. We thus adopt the cross-attention version for the local reconstruction module.

\textbf{Hyper-Parameters Search of LR:}
We test different hyper-parameters of the local reconstruction module in Fig.~\ref{exp_mlm_ablation}. Our method achieves the best performance when the ratio of random masking is 0.15, the number of attention blocks is 3, and the number of MLP is 4. The local reconstruction module aims to reconstruct the masked textual words conditioned on the reference. Setting the masking ratio as 0.15 leads to an appropriate difficulty for the subsequent reconstruction procedure, and thus gets the best performance.

\begin{figure*}
	\begin{center}
		\includegraphics[width=1\linewidth]{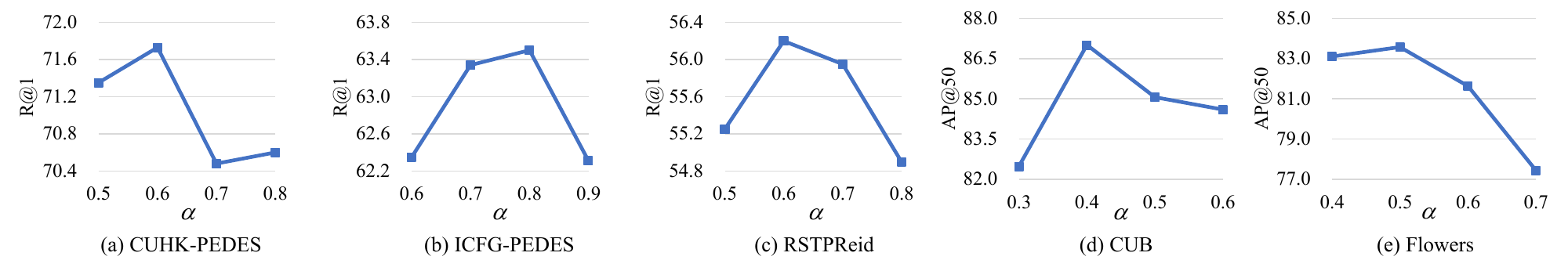}
	\end{center}
	\caption{{Ablation studies of the lower bound $\alpha$ in Eq.~(\ref{loss_func}) across different datasets. The upper bound $\beta$ is set to $\alpha-0.2$ following ViTAA~\cite{wang2020vitaa}.}}
	\label{exp_alpha_ablation}
\end{figure*}

\textbf{ Hyper-Parameters $\alpha$ and $\beta$ in Eq.~(\ref{loss_func}):}
We test the value of the lower bound $\alpha$ for positive similarity and upper bound $\beta=\alpha-0.2$ for negative similarity on different datasets in Fig.~\ref{exp_alpha_ablation}. The optimal value of $\alpha$ for each dataset may be slightly different. However, the optimal values range from 0.4 to 0.8 across different datasets, which might simplify the hyper-parameter tuning.

\begin{table}
    \caption{Experiments on domain generalization setting with ResNet50 backbone.
    ``BOT~(Image)'' represents a strong image-based retrieval baseline~\cite{Luo_2019_CVPR_Workshops}. ``MMRef~(Image)'' denotes our MMRef trained only with images. ``MMRef~(Text+Image)'' means training with texts and images. CUHK$\dagger$ denotes the subset of CUHK-PEDES which excludes Market and MSMT17. }
   \label{domain_gen}
   \small
   \centering
   \setlength{\tabcolsep}{6pt}
   \begin{tabular}{c|c|c|cc}
   \hline
   Source & Target & Method & R@1 & mAP\\ \hline\hline
    CUHK$\dagger$ & CUHK & BOT~(Image) & 92.5 & 59.2\\
    CUHK$\dagger$ & CUHK & MMRef~(Image) & \textbf{93.7} & \textbf{60.4}\\
    CUHK$\dagger$ & CUHK & MMRef~(Text+Image) & 92.29 & 58.74\\ \hline
    CUHK$\dagger$ & Market & BOT~(Image) & 50.8 & 29.9\\
    CUHK$\dagger$ & Market & MMRef~(Image) & 66.0 & 43.9\\
    CUHK$\dagger$ & Market & MMRef~(Text+Image) & \textbf{71.2} & \textbf{48.0}\\ \hline
    CUHK$\dagger$ & MSMT17 & BOT~(Image) & 13.5 & 5.2 \\ 
    CUHK$\dagger$ & MSMT17 & MMRef~(Image) & 15.9 & 5.9 \\ 
    CUHK$\dagger$ & MSMT17 & MMRef~(Text+Image) & \textbf{32.4} & \textbf{11.9} \\ \hline
   \end{tabular}
  
\end{table}

\subsection{Text-Assisted Domain Generalization}

To validate that visual representations aligned with text descriptions can boost the domain generalization capability, we compare
a strong image-based person re-identification baseline ``BOT''~\cite{Luo_2019_CVPR_Workshops}, ``MMRef(Image)'' trained only with images,  and ``MMRef~(Text+Image)'' trained with both texts and images in the domain generalization setting. In this setting, each method is first trained on the source dataset CUHK$\dagger$ and then tested on different target datasets.

\begin{figure}[t]
	\begin{center}
		\includegraphics[width=0.8\linewidth]{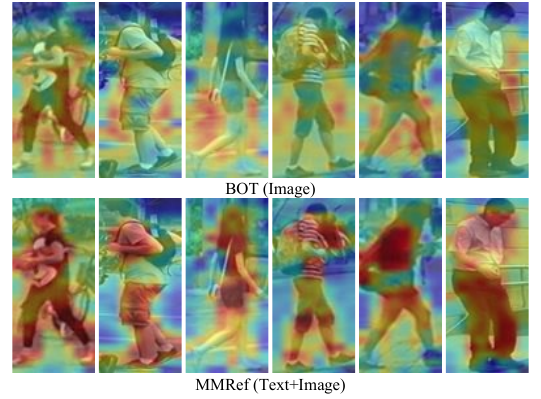}
	\end{center}
	\caption{
		{Attention visualization of visual features extracted by BOT(Image) and our MMRef(Text+Image) on the unseen Market dataset.} }
	\label{domain_attention}
\end{figure}

As shown in Tab.~\ref{domain_gen}, compared with MMRef~(Image) and MMRef~(Text+Image), the strong baseline BOT~\cite{Luo_2019_CVPR_Workshops} achieves comparable performance on CUHK. 
However, BOT trained on CUHK$\dagger$ achieves the lowest performance on unseen domains, \textit{i.e.}, Market and MSMT17, showing that the BOT model trained with images has poor domain generalization capability. 

\begin{figure}
	\begin{center}
		\includegraphics[width=0.8\linewidth]{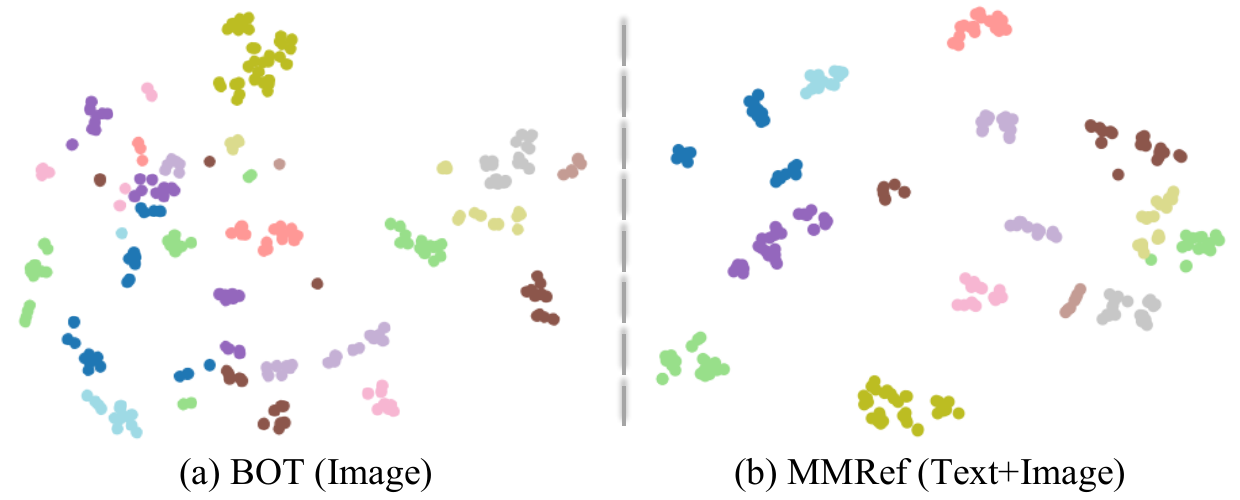}
	\end{center}
	\caption{
		{t-SNE visualization of visual features extracted from MSMT17.} }
	\label{domain_img}
\end{figure}
We observe a notable performance gap between BOT and MMRef~(Image) in the Market dataset. It is reasonable because ResNet50 in MMRef~(Image) is initialized by pre-trained weights of CLIP while BOT~\cite{Luo_2019_CVPR_Workshops} uses the weights pre-trained on ImageNet. It shows that the visual backbone in MMRef~(Image) also benefits from textual knowledge. 
When we directly align the visual representations with textual descriptions in MMRef~(Text+Image), the performance on unseen domains can be consistently improved.

As illustrated in Fig.~\ref{domain_attention}, the visual feature extracted by BOT tends to focus on noisy backgrounds, when evaluated in the unseen domain. In contrast, our MMRef pays more attention to the person body areas. This illustrates the effectiveness of aligning visual features and textual descriptions. Given that backgrounds can vary significantly across images and domains, focusing on meaningful person body areas leads to more robust visual features, and enhances the generalization capability in unseen domains.

The t-SNE visualization of features from MSMT17 extracted by BOT and our MMRef~(Text+Image) is shown in Fig.~\ref{domain_img}. We can find that the features of the same person extracted by our MMRef(Text+Image) are more concentrated while the features belonging to different persons are easier to distinguish.

Based on those visualizations and the results in Table~\ref{domain_gen}, we conclude that aligning the images with textual descriptions can boost the domain generalization capability of visual representations in conventional image-based person retrieval.

\begin{figure*}
	\begin{center}
		\includegraphics[width=0.8\linewidth]{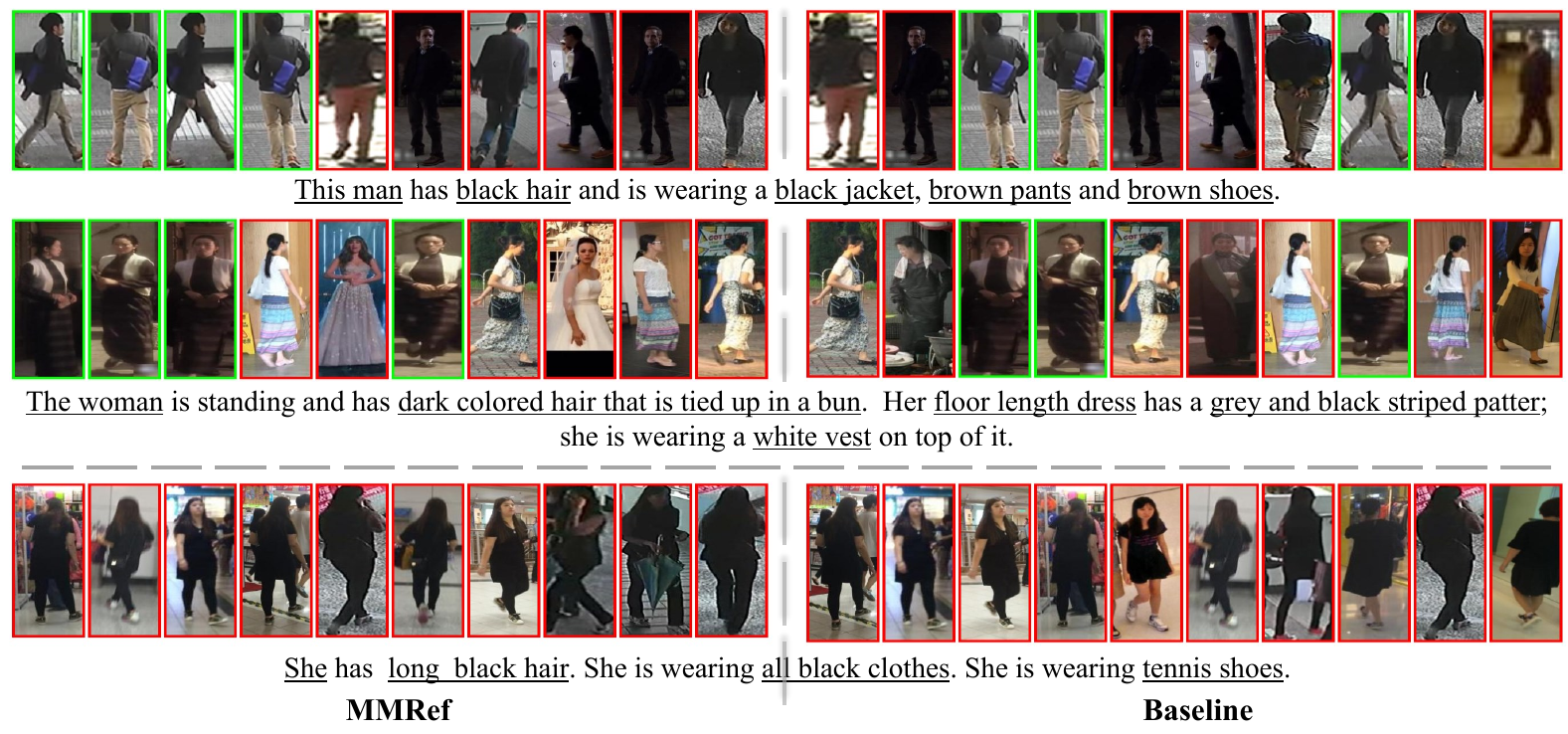}
	\end{center}
	\caption{{Examples of top-10 retrieval results on CUHK-PEDES test set.} Green boxes denote true positives, while the red boxes mean false negatives. }
	\label{case_analysis}
\end{figure*}

\begin{figure*}
	\begin{center}
		\includegraphics[width=0.8\linewidth]{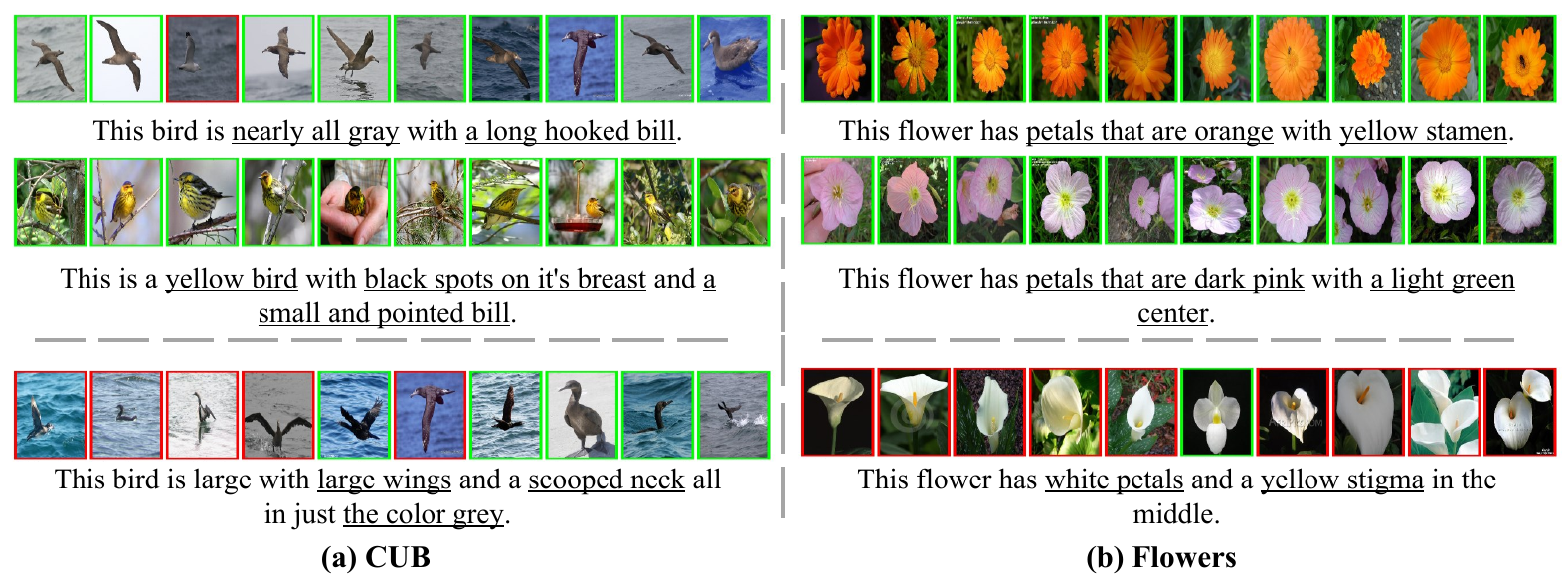}
	\end{center}
	\caption{{Visualization of top-10 retrieval results on CUB and Flowers test set.} Green boxes denote true positives, while the red boxes mean false negatives. }
	\label{case_analysis2}
\end{figure*}

\subsection{Analysis of Retrieval Results:} 
We compare the top-10 retrieval results of our proposed MMRef and the baseline in Fig.~\ref{case_analysis}.
As shown in the first three rows of Fig.~\ref{case_analysis}, our MMRef obtains more accurate retrieval results than the baseline, which shows that our method enables the model to learn more discriminative uni-modal representations.
In the last row, both our MMRef and the baseline fail to retrieve the correct images of the target person. The incorrect retrieved images correspond to all textual attributes in the textual query, such as ``long black hair'',  ``all black clothes'' and ``tennis shoes''. The failure of retrieval mainly results from the non-discriminative textual query. It indicates one of the limitations of text-based person ReID is that textual query may not be sufficiently discriminative. When the text query describes the image at a coarse level, there may exist several corresponding images from different persons. The clarity of text query has a huge influence on retrieval accuracy. 
In Fig.~\ref{case_analysis2}, we illustrate the retrieval results on CUB and Flowers. Our MMRef can successfully retrieve the target images in most cases if the text query can clearly describe the discriminative details. In the last row, failures occur when the text query is ambiguous.
\section{Conclusion}
This paper proposes a novel multi-modal reference learning framework, named MMRef, to mitigate the effects of inaccurate and incomplete text annotations. Specifically, we fuse multi-modal details of an object to construct a multi-modal reference with our proposed global fusion module and local reconstruction module.
As a comprehensive representation of an object, the reference, in turn, facilitates learning better uni-modal visual and textual representations. The multi-modal references constructed in the training stage can also refine the retrieval results with our reference-based refinement method. Extensive experiments on fine-grained text-to-image retrieval datasets show that our method outperforms existing approaches by notable margins. Meanwhile, our experiments show that aligning images with text descriptions can effectively boost the domain generalization ability of visual features for fine-grained image-based person retrieval.

\bibliographystyle{IEEEtran}  
\bibliography{reference} 

\begin{IEEEbiography}[{\includegraphics[width=1in,height=1.25in,clip,keepaspectratio]{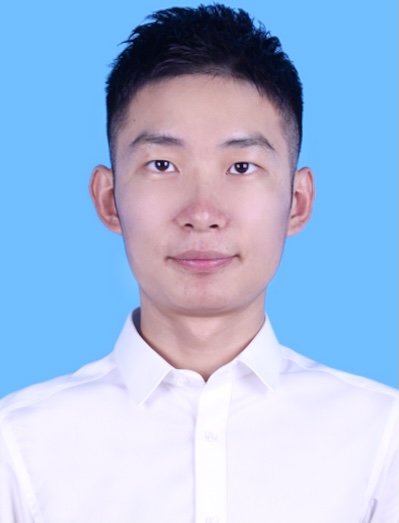}}]{Zehong Ma} received the B.S. degree from Northwestern Polytechnical University, Xi'an, China, in 2022. He is currently pursuing a Ph.D degree at the School of Computer Science, Peking University, Beijing, China. His current research interests are multi-modal representation learning, with a focus on multi-modal retrieval and open-vocabulary recognition.
\end{IEEEbiography}

\begin{IEEEbiography}[{\includegraphics[width=1in,height=1.25in,clip,keepaspectratio]{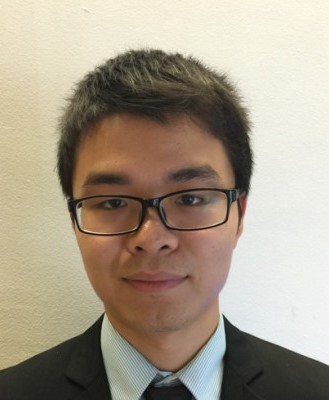}}]{Hao Chen}
received the Ph.D. degree in computer science from INRIA in 2022. 
He is currently a Post-Doctoral Researcher at Peking University. His research
interests include fine-grained image understanding, unsupervised learning and incremental learning. 
\end{IEEEbiography}

\begin{IEEEbiography}[{\includegraphics[width=1in,height=1.25in,clip,keepaspectratio]{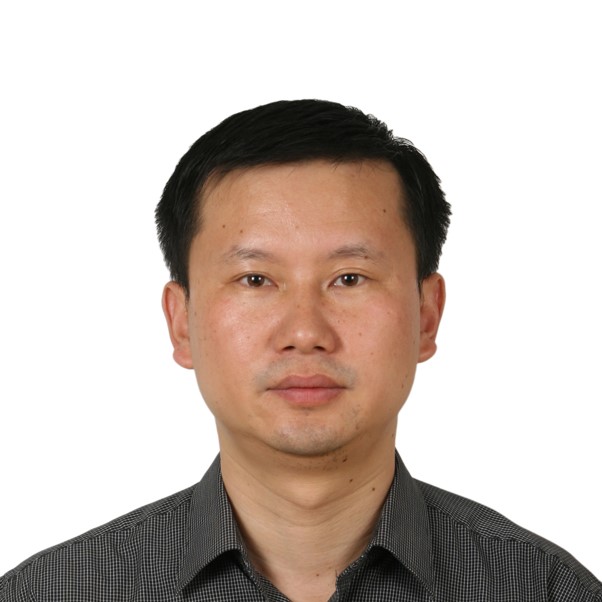}}]{Wei Zeng} is currently a researcher in the School of Computer Science at Peking University, P. R. China. 
He was a senior researcher at NEC Laboratories, China from 2005 to 2015. He worked as a visiting scholar at Stanford University in 2012. He received his PhD degree in Computer Science and Engineering from the Harbin Institute of Technology, in 2005. 
His research interests include computer vision (object detection, segmentation), artificial intelligence (AI computing platform), and media analysis (video analysis, retrieval). 
He is the author or coauthor of over 40 refereed journals and conference papers. He was the reviewer of ECCV2022, CVPR2022, ICPR2020, BigMM2020, etc.
\end{IEEEbiography}

\begin{IEEEbiography}[{\includegraphics[width=1in,height=1.25in,clip,keepaspectratio]{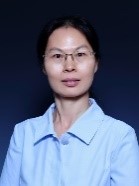}}]
{Limin Su} received the Ph.D. degree in Control Theory and Control Engineering from Beijing Institute of Technology in 2003. She serves as an associate professor, and the Head of the Department of Data Science at Beijing Union University. Her current research interests include machine learning, Big Data technology and signals and information processing.
\end{IEEEbiography}

\begin{IEEEbiography}[{\includegraphics[width=1in,height=1.25in,clip,keepaspectratio]{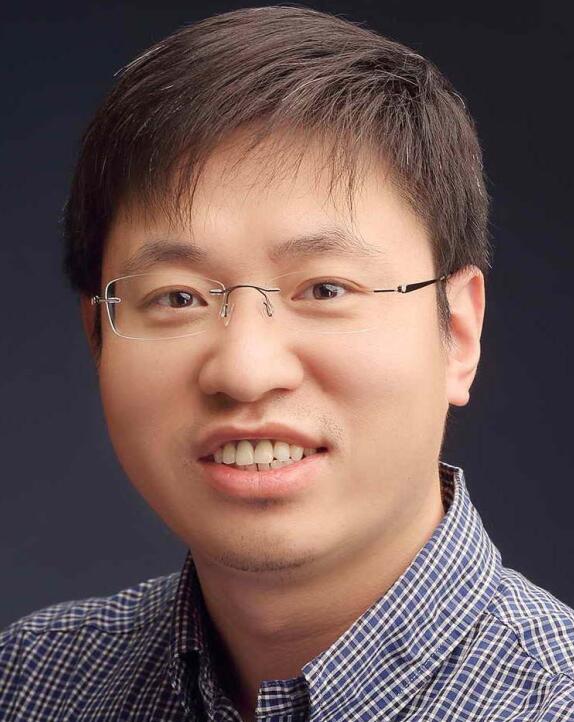}}]{Shiliang Zhang} (Senior Member, IEEE) received the PhD degree in computer science from the Institute of Computing Technology, Chinese Academy of Sciences. He was a post-doctoral scientist with the NEC Laboratories America and a post-doctoral research fellow with The University of Texas at San Antonio. He is currently an associate professor with Tenure with the School of Computer Science, Peking University. His research interests include large-scale image retrieval, visual perception, and computer vision. He has authored or co-authored more than 100 papers in journals and conferences, including International Journal of Computer Vision, IEEE Transactions on Pattern Analysis and Machine Intelligence, IEEE Trans. on Image Processing, IEEE Transactions on Neural Networks and Learning Systems, IEEE Trans. on Multimedia, ACM Multimedia, ICCV, CVPR, ECCV, AAAI, IJCAI, etc. He was a recipient of the Okawa Foundation Research Award, Outstanding Doctoral Dissertation Awards from the Chinese Academy of Sciences and Chinese Computer Federation, the NEC Laboratories America Spot Recognition Award, the NVidia Pioneering Research Award, and the Microsoft Research Fellowship, etc. He was a recipient of the Top 10\% Paper Award with the IEEE MMSP. He served as the associate editor (AE) of Computer Vision and Image Understanding (CVIU) and IET Computer Vision, guest editor of ACM Transactions on Multimedia Computing, Communications, and Applications, and area chair or Senior Program Committee of ICCV, CVPR, AAAI, ICPR, and VCIP. His research is supported by The National Key Research and Development Program of China, Natural Science Foundation of China, Beijing Natural Science Foundation, and Microsoft Research, etc.
\end{IEEEbiography}

\vfill

\end{document}